\documentclass[10pt,twocolumn,letterpaper]{article}
\usepackage[T1]{fontenc}
\usepackage[utf8]{inputenc}
\usepackage{authblk}

\usepackage{iccv}
\usepackage{times}
\usepackage{epsfig}
\usepackage{graphicx}
\usepackage{amsmath}
\usepackage{amssymb}

\usepackage{lipsum}
\usepackage{placeins}
\usepackage{booktabs} % For formal tables
\usepackage{subfigure}
\usepackage{multirow}
\usepackage{ulem}
\usepackage{pifont}% http://ctan.org/pkg/pifont
\usepackage{hhline}
\usepackage{nopageno}

\usepackage[ruled]{algorithm2e} % For algorithms

\usepackage[breaklinks=true,bookmarks=false]{hyperref}

\iccvfinalcopy % *** Uncomment this line for the final submission

\def\iccvPaperID{****} % *** Enter the ICCV Paper ID here

\begin{document}

%%%%%%%%% TITLE
\title{VrR-VG: Refocusing Visually-Relevant Relationships\thanks{This work was performed at JD AI Research.}}

\author[1,2]{Yuanzhi Liang}%\footnotemark[1]\thanks{This work was done when Yuanzhi Liang was a research intern in JD AI research.}
\author[2]{Yalong Bai}
\author[2]{Wei Zhang}
\author[1]{Xueming Qian}
\author[1]{Li Zhu}
\author[2]{Tao Mei}
\affil[1]{Xi'an Jiaotong University} \affil[2]{JD AI Research, Beijing, China}
\affil[ ]{\tt\small {liangyzh13@stu.xjtu.edu.cn} {ylbai@outlook.com} {wzhang.cu@gmail.com} {\{qianxm, zhuli\}@mail.xjtu.edu.cn} {tmei@live.com} }

%\author{Yuanzhi Liang\\
%Xi'an Jiaotong University \\
% For a paper whose authors are all at the same institution,
% omit the following lines up until the closing ``}''.
% Additional authors and addresses can be added with ``\and'',
% just like the second author.
% To save space, use either the email address or home page, not both
%\and
%Yalong Bai\\
%%JD AI Research\\
%{\tt\small liangyzh13@stu.xjtu.edu.cn \tt\small ylbai@outlook.com}
%}

\maketitle
%\thispagestyle{empty}

%%%%%%%%% ABSTRACT
\vspace{-.1in}
\begin{abstract}
  Relationships encode the interactions among individual instances, and play a critical role in deep visual scene understanding. Suffering from the high predictability with non-visual information, existing methods tend to fit the statistical bias rather than ``learning'' to ``infer'' the relationships from images.
  To encourage further development in visual relationships, we propose a novel method to automatically mine more valuable relationships by pruning visually-irrelevant ones. We construct a new scene-graph dataset named Visually-Relevant Relationships Dataset (VrR-VG) based on Visual Genome.
  Compared with existing datasets, the performance gap between learnable and statistical method is more significant in VrR-VG, and frequency-based analysis does not work anymore. Moreover, we propose to learn a relationship-aware representation by jointly considering instances, attributes and relationships. By applying the representation-aware feature learned on VrR-VG, the performances of image captioning and visual question answering are systematically improved with a large margin, which demonstrates the gain of our dataset and the features embedding schema. VrR-VG is available via \url{http://vrr-vg.com/}.%\href{http://vrr-vg.com/}{http://vrr-vg.com/}.
\end{abstract}

%%%%%%%%% BODY TEXT
%\blfootnote{This work was done when Yuanzhi Liang was a research intern in JD AI research.}
%\vspace{-.3in}
\section{Introduction}
%\label{sec:intro}

Although visual perception tasks (e.g., classification, detection) have witnessed great advancement in the past decade, visual cognition tasks (e.g., image captioning, question answering) are still limited due to the difficulty of reasoning~\cite{visual-genome}. Existing vision tasks are mostly based on individual objects analysis. However, a natural image usually consists of multiple instances in a scene, and most of them are related in some ways. To fully comprehend a visual image, a holistic view is required to understand the relationships and interactions among object instances. 

\begin{figure}[!t]
    \centering
    \includegraphics[page=1,width=0.45\textwidth]{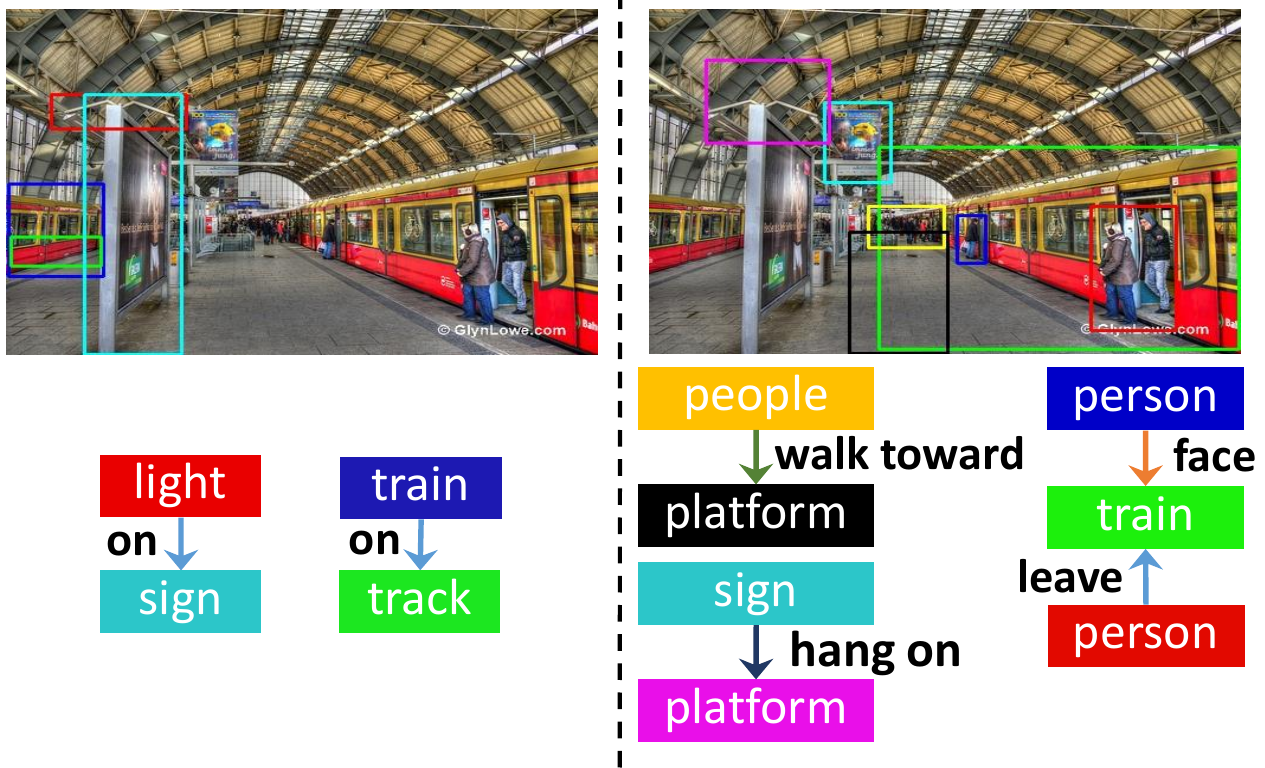}
    \caption{Example scene graphs in VG150 (left) and VrR-VG (right, ours). More visually-relevant relationships are included in VrR-VG.}% More examples in Supplementary xx.}
    \label{scene_graph_eg_intro}
%\vskip -1em
\end{figure}

Visual relationships \cite{visual-relationshp-detection-with-language-priors, detecting-visual-relationships-with-deep-relational-networks,iterative-message-passing, neural-motifs, large-scale-visual-relationship-understanding}, which encode the interplay between individual instances, become the indispensable factor for visual cognitive tasks such as image captioning \cite{yao-ting-caption}, visual question answering (VQA) \cite{gcn-vqa}.
% define visual relationship: scene graph and triplets
In existing literature, visual relationships are mostly represented as a \textit{scene graph} (Fig.~\ref{scene_graph_eg_intro}): a node represents a specific instance (either as subject or object), and an edge encodes the \textit{relation label} (\textit{r}) between a \textit{subject} (\textit{s}) and an \textit{object} (\textit{o}). Equivalently, a scene graph can also be represented as a set of triplets $\left \langle  \textit{s, r, o} \right \rangle$. Recently, extensive research efforts \cite{iterative-message-passing,neural-motifs,pixels-to-graphs-by-associative-embedding,graph-rcnn-for-scene-graph-generation} are conducted on \textit{scene graph generation}, which aims to extract the scene graph from an image (Fig.~\ref{scene_graph_eg_intro}). Essentially, scene graph generation bridges the gap between visual perception and high-level cognition. 
% For simplicity of notation, we adopt the triplets notation hereafter in this paper. 

\iffalse
\begin{figure}
    \centering
    \includegraphics[page=1,width=0.95\linewidth]{figures.pdf}
    \label{graph_example_intro}
    \caption{An overview of the scene graph generation task. With different conditions of inputs, the scene graph generation task can be treated as Scene Graph Detection (SGDet), Scene Graph Classification (SGCls), Predicates classification (PredCls), and Predicate detection (PredDet).
    %The upper figures show a ground truth scene graph containing several entities, such as man, horse, tail, etc. These entities are localized with bounding boxes. The relationships are represented as the edges connecting those entities. The relationships, combined with entities, are expressed as multiple triplets or organized as a scene graph structure. The lower figure shows four different settings of scene graph generation tasks. Given image with different kinds of auxiliary information, the final goal of scene graph generation task is to generate a comprehensive scene graph corresponding to the input image.
    }
\end{figure}
\fi

\begin{figure*}[htbp!]
    \centering
    \includegraphics[page=10,width=0.85\linewidth]{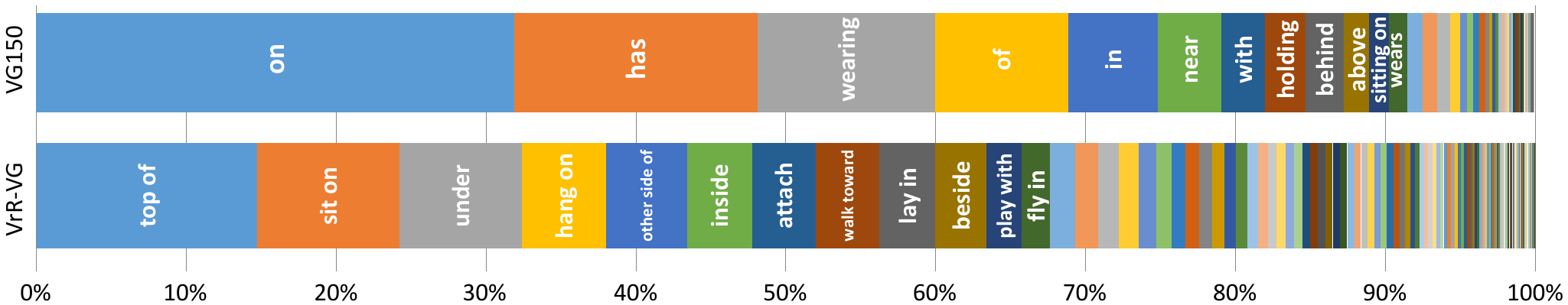}
    \caption{Distribution of relation labels in VG150 (top) and VrR-VG (bottom). Our VrR-VG is more diverse and balanced than VG150.}
    \label{dataset_rela_distribution}
\end{figure*}

\iffalse
\begin{figure*}[!t]
    \centering
    \includegraphics[page=1,width=\linewidth]{datasetstatistic.pdf}
    \caption{Histogram of relationship labels in VG150, ranked by label frequency. The most frequent 16 relation labels (in bold) take up 93.6\% in the dataset.}
    \label{distribution4into}
\end{figure*}
\fi

% dataset & task
Among the datasets \cite{VisualPhrases,visual-genome,visual-relationshp-detection-with-language-priors,open-image, unrel} adopted in visual relationship, Visual Genome (VG) \cite{visual-genome} provides the largest set of relationship annotations, offering large-scale (2.3 million relationships) and dense (21 relationships per image) relationship annotations. However, the relationships in VG are heavily noisy, biased and duplicated, since it was automatically extracted from image captions. VG150\footnote{We call it ``VG150" to distinguish from the original VG dataset \cite{visual-genome}.} \cite{iterative-message-passing}, the most popular split derived from VG, is constructed by only keeping the most frequent 150 object categories and 50 relation labels in VG. In existing literature, VG150 serves as the most widely adopted benchmark on scene graph generation \cite{neural-motifs,iterative-message-passing, graph-rcnn-for-scene-graph-generation, add4vg150, pixels-to-graphs-by-associative-embedding, add4vg150_2}, but was seldomly adopted on cognitive tasks such as captioning and VQA.

% 现在relationship不work，no real effect. need some evidence on the problems of visual relationships.
%There have been intensive research efforts on visual relationships \cite{}, and most of which are based on the popular VG150 dataset.
% Although visual relationships are important ingredients for high-level deep image understanding, it does not show too much improvement as expected on a range of tasks. Surprisingly, visual relationship was found to have negative effect on image captioning and visual question answering tasks \cite{}. This phenomenon suggests that current research on visual relationship is not properly posed, at least is not designed for high-level image understanding tasks such as image captioning. 

% problems in current datasets
% As pointed out in \cite{neural-motifs}, current relationship detection can be well addressed by

% 问题在于不是真正的VISUAL relationship  X
%However, there are still two major defects in existing datasets including VG150. First, no high-level tasks benefit from visual relationship, to the best of our knowledge. Second, 
%Though popular on scene graph generation, VG150 was never adopted on cognitive tasks such as captioning and VQA. 
Based on our study, there are still several problems in current visual relationship datasets: visual relationships are actually not that \textit{``visual"}. That is, a large portion of relationships are visually irrelevant. 
% Taking VG150 as an example,
% We found that the relationship data in VG150 has a severe defect, which cause the relation representation heavily relies on non-visual information (i.e., bias of relationship distribution, language prior of labels), rather than visual information.
%
1) Some \textit{spatial relationships} (e.g., ``on'', ``of'', ``in'') are less visually informative. 
As shown in Fig.~\ref{dataset_rela_distribution}, spatial relationships take up a substantial proportion in VG150. For example, ``on'' takes 31.9\% in all relation labels. However, some spatial relationships can be easily inferred merely based on the bounding box locations of \textit{s} and \textit{o}, without even accessing the visual content.
%According to \cite{neural-motifs}, given a list of labeled boxes pairs with edges in the ground truth, the prediction accuracy of relationships reaches 96.0\% in terms of R@50, 98.4\% in terms of R@100\footnote{R@$N$: the fraction of times the correct relationship is predicted in the top-$N$ predictions.} in VG150. \textcolor{blue}{The results indicate that the relation task degrees to single object detection task, and achieves excessive high performance when given paired detection ground truth.}
2) Large portion of \textit{low diversity} relation labels gives rise to frequency analysis. Some relationships (e.g., ``wear'', ``ride'', ``has'') can be roughly estimated only based on language priors or statistical measures, without looking at the visual image. As shown in Fig.~\ref{object_bias_eg}, given ``\textit{s=man}'' and ``\textit{o=nose}'', 95.8\% of \textit{r} is ``\textit{has}''.
%Similarly, when taking ``man" as subject and ``jacket" as object, 69.32\% of labeled relationship is ``wearing". 
Results in \cite{neural-motifs} also show that simple frequency-counting achieves decent results in many metrics of scene graph generation, which indicates many relation labels in VG150 can be predicted by non-visual factors.
% outperforms several learning-based methods, indicating many relation labels can be predicted according to non-visual factors.
% Results in \cite{neural-motifs} show that R@50\footnote{The fraction of times the correct relationship is predicted in the top-50 predictions.} reaches 96.0\%, given the paired ground-truth detection. By contrast, R@50 is only 27.2\% without any detection ground truth.
% The high predictability using statistical priors is also contrary to the original intention of the visual relationship.
% The lack of diversity leads current task to frequency symptom, which needs little visual analysis.
%Current studies on visual relationships are not properly posed, and visual information has not been fully explored for visual relationship. 
Due to these problems, cognitive tasks (e.g., image captioning, VQA) can hardly benefit from relationships learned from current datasets. To the best of our knowledge, no cognitive tasks have benefited from current visual relationship dataset so far, except a few \cite{yao-ting-caption, gcn-vqa} not learning from visual relationship datasets. 
These phenomenons suggest that current datasets on the visual relationship are quite limited.  %, at least is not designed for high-level image understanding tasks such as image captioning.

\iffalse
\begin{figure}[!t]
    \begin{minipage}[t]{\linewidth}
        \centering
        \includegraphics[page=4,width=0.8\linewidth]{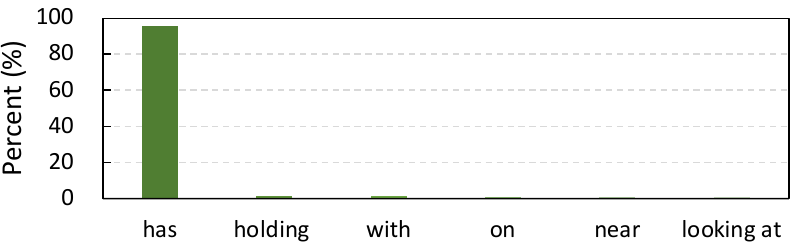}\\
        (a) \textit{s = man, o = nose}
    \end{minipage}
    \hfill
    \begin{minipage}[t]{\linewidth}
        \centering
        \includegraphics[page=5,width=0.8\linewidth]{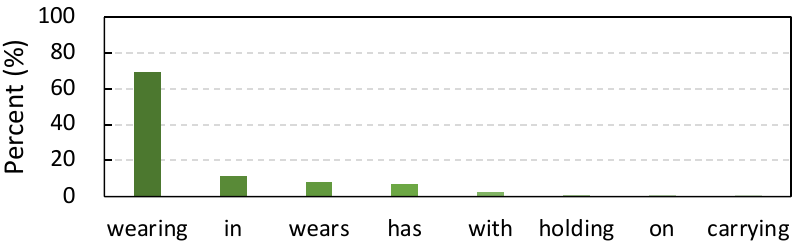}\\
        (b) \textit{s = man, o = jacket}
    \end{minipage}
    \caption{Distribution of relation labels when (a) ``s = man, o = nose", and (b) ``s = man, o = jacket". Low diversity of relation labels is observed in VG150.}
    \label{object_bias_eg}
\end{figure}
\fi
\begin{figure}[!t]
    \centering
    \includegraphics[page=1,width=0.85\linewidth]{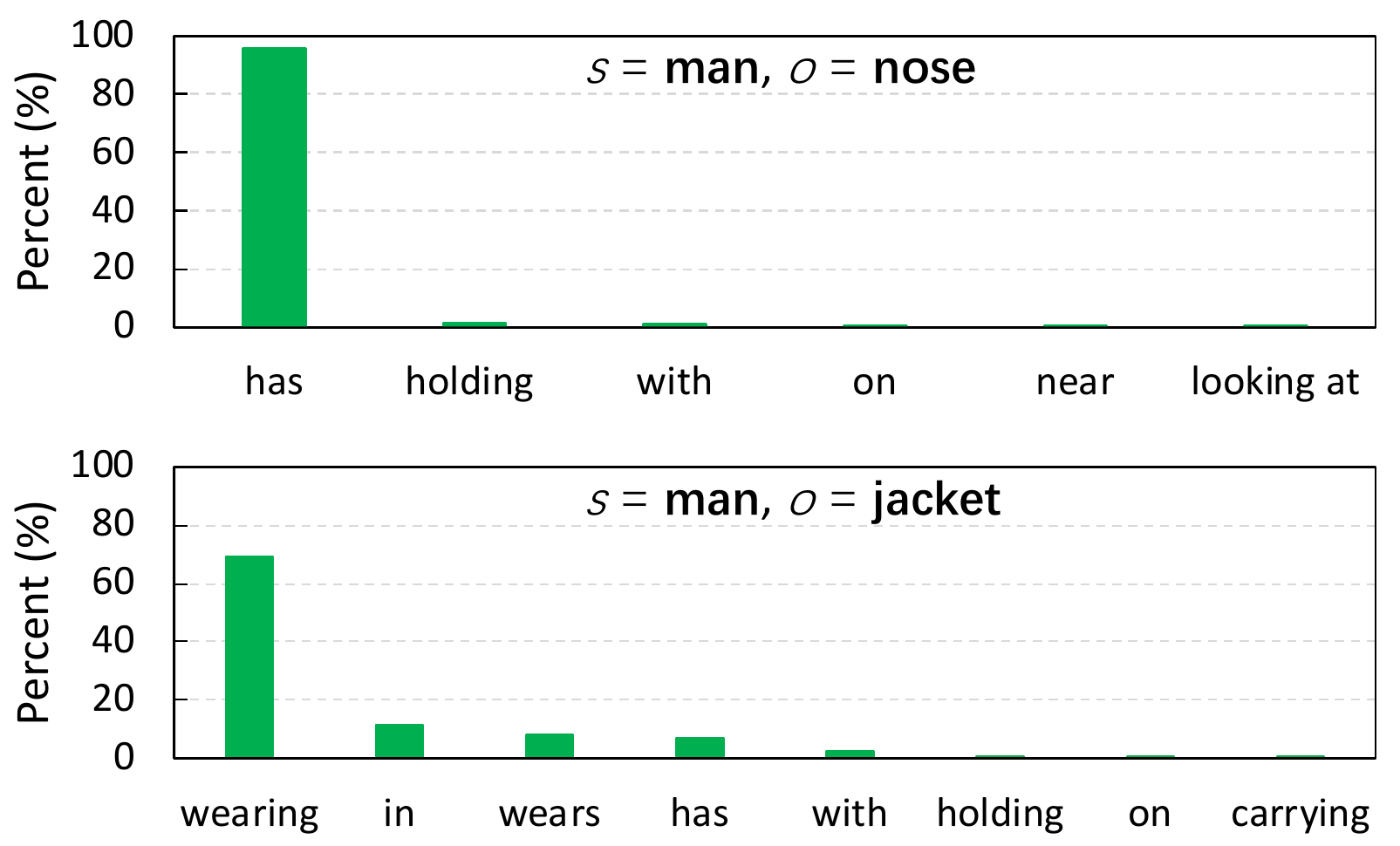}
    \caption{Distribution of relation labels when ``s = man, o = nose", and ``s = man, o = jacket". Low diversity of relation labels is observed in VG150.}
    \label{object_bias_eg}
\vskip -1em
\end{figure}

\begin{figure*}[htb]
    \begin{minipage}[t]{.45\textwidth}
        \centering
        \includegraphics[page=6,width=0.75\linewidth]{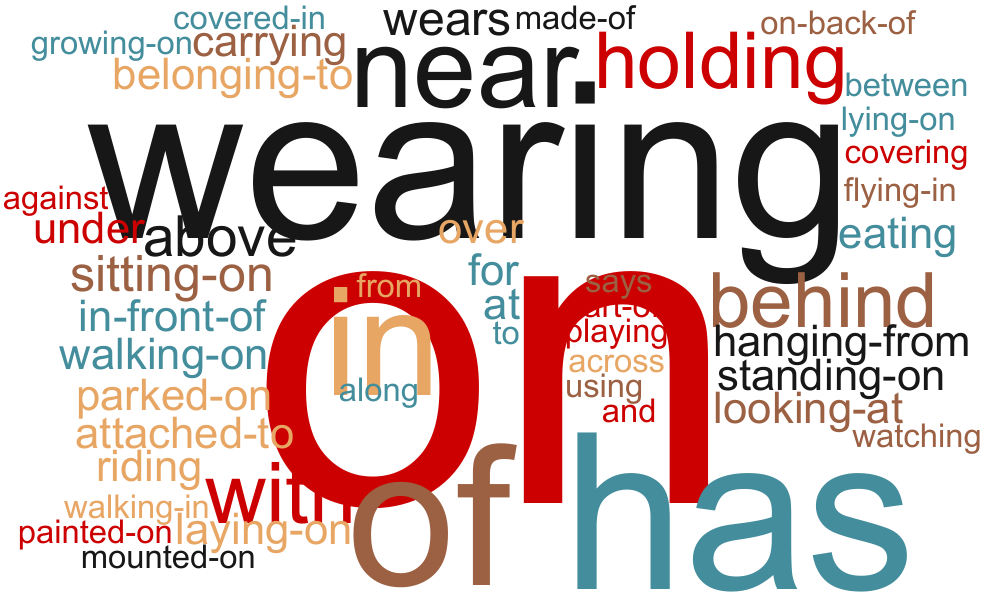}
    \end{minipage}
    %\\
%	\\
	\hfill
    \begin{minipage}[t]{.45\textwidth}
        \centering
        \includegraphics[page=7,width=0.8\linewidth]{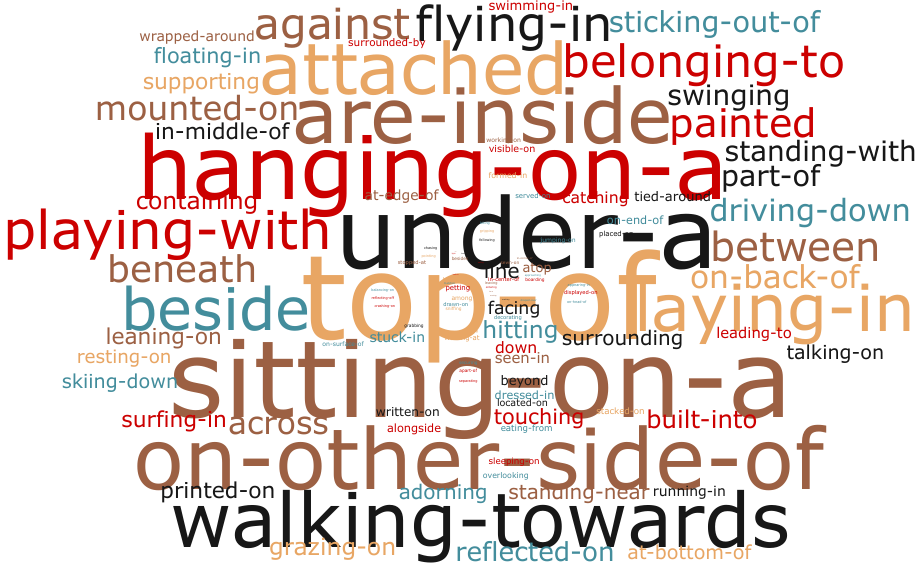}
    \end{minipage}
    \caption{Tag cloud visualization for VG150 \cite{iterative-message-passing,neural-motifs} (left) and VrR-VG (right, ours). VrR-VG covers more visually-relevant relationships.}
    \label{tag_cloud}
\end{figure*}

In this paper, we propose a novel method to automatically identify visually-relevant relationships and construct a new data split named Visually-relevant Relationships (VrR-VG) %\footnote{The dataset is available at http://vrr-vg.com/.}
based on the original VG. Specifically, a tiny visual discriminator network (VD-Net) is carefully designed to learn the notion of visually-relevant. To exploit the full capacity of VrR-VG on cognitive tasks, we also propose a joint learning method for relationship-aware representation learning.
We show that VD-Net is effective in pruning visually-irrelevant relationships from the large corpus. 
% Fig.~\ref{scene_graph_eg_intro} shows an example comparison of VG150 and VrR-VG.
Compared to VG150, VrR-VG focuses more on visually-relevant relations (Fig.~\ref{scene_graph_eg_intro} and \ref{tag_cloud}), and is more balanced in label distribution (Fig.~\ref{dataset_rela_distribution}). 
Our experiments show that non-visual based methods no longer work well on VrR-VG.  %More importantly, visual relationship features learned on VrR-VG are much more effective than those from previous datasets. 
% To demonstrate the properties of VrR-VG, we report the performances comparison in scene graph generation.
% The experiments show significant performance decrease when adopting our data split. We also prove that barely frequency analysis no longer works in VrR-VG. 
%The experiments show that the performance of scene graph generation in VrR-VG is heavily dependent on visual information extracted from images, and the barely frequency analysis no longer works in VrR-VG.
% The experiments show the significant gap of performance between frequency based and learnable method in our dataset, and the barely frequency analysis no longer works in relation estimation.
More importantly, relationship-aware features learned on VrR-VG show more promising results in cognition tasks such as VQA and image captioning. This also indicates that more valuable visual relationships are included in our dataset. %All these results provide further proof of the defects in previous datasets. %and also expose insufficiency in current relationships research works. 
The new dataset (VrR-VG) and our pre-trained relationship features will be released to the community to facilitate further researches on scene graph understanding and high-level cognitive tasks.
The main contributions of this paper are summarized as follows:

1. A new dataset VrR-VG is constructed to highlight visually-relevant relationships. For this purpose, we also propose a novel visual discriminator to learn the notion of visually-relevant.

2. We propose a relationship-aware feature learning schema for incorporating object instances and their relationships into one feature vector. Objects location / category / attribute as well as their relations are jointly considered, such that semantics and their relations are jointly modeled.

3. Better results on visual cognitive tasks (VQA and image captioning) further verifies the effectiveness of our VrR-VG dataset as well as the relationship-aware feature learning schema.

\iffalse
\begin{enumerate}
%\setlength\itemsep{0em}
    \item A new dataset VrR-VG is constructed to highlight visually-relevant relationships. For this purpose, we also propose a novel visual discriminator to learn the notion of visually-relevant.
    %Both of the Spatial Relationships and Statistically Biased Relationships are detected using this method. Complex and visually-relevant relationships are reserved. 
    %Our dataset has a comparable scale with previous datasets but more visually-relevant relationships are included. 
    \item We propose a relationship-aware feature learning schema for incorporating object instances and their relationships into one feature vector. Objects location / category / attribute as well as their relations are jointly considered, such that semantics and their relations are jointly modeled.
    \item Better results on visual cognitive tasks (VQA and image captioning) further verifies the effectiveness of our VrR-VG dataset as well as the relationship-aware feature learning schema.
    %. Our proposed dataset and relationship feature learning schema show significant performance boosts on visual cognitive tasks. 
\end{enumerate}
\fi

\section{Related Work}
%To have a holistic interpretation in the image, the relationships of entities are essential. It is not enough to understand image as several isolated objects by detection. As the bond of entities, 
%%rather than focusing on isolated objects in detection, 
%relationships provide a new view to understand the interactions between entities and try to figure out the semantic information from multiple entities in images. For going deeper into this topic, many datasets provide relationships annotation, and many works show the tremendous potential of relationships. 
%In this section, we describe related works in dataset and methods in relationship research.
%\subsection{Datasets in Relationships}

\textbf{Visual relationship datasets:} We summarize some datasets in visual relationship in Table~\ref{dataset_table}. Visual phrase dataset \cite{VisualPhrases} focus on relation phrase recognition and detection, which contains 8 object categories from Pascal VOC2008 \cite{pascal-voc2008} and 17 relation phrases with 9 different relationships. Scene Graph dataset~\cite{image-retrieval-using-scene-graphs} mainly explores the ability of image retrieval by scene graph. The VRD dataset \cite{visual-relationshp-detection-with-language-priors} intends to benchmark the scene graph generation.
%, and contains 37,993 relation triplets
Open Images \cite{open-image} provides the largest amount of images for object detection and also presents a challenging task for relationship detection. PIC \cite{PIC} proposes a segmentation task in the context of visual relationship. 

Visual Genome (VG) \cite{visual-genome} has the maximum amount of relation triplets with the most diverse object categories and relation labels in all listed datasets. %As shown in Table~\ref{dataset_table}, millions of instance labels, relation triplets are contained in VG. 
However, the relations in VG contain lots of noises and duplications. Thus VG150~\cite{iterative-message-passing} is constructed by pre-processing VG by label frequency. However, most high-frequency relationships are visually-irrelevant as we mentioned before. 

%, and the massive relationships annotations made VG become one of the most popular datasets in many research fields, like object detection %\cite{vg-detection1,vg-detection2,vg-detection3,vg-detection4}, VQA \cite{vg-vqa0,vg-vqa1,vg-vqa2, vg-vqa3,vg-vqa4}, caption \cite{vg-cap0,vg-cap1,vg-cap2,scst,lstm-a},  relation detection \cite{visual-relationshp-detection-with-language-priors, visual-translation-embedding-network-for-visual-relation-detection, detecting-visual-relationships-with-deep-relational-networks, visual-translation-embedding-network-for-visual-relation-detection, deep-variation-structured-reinforcement-learning-for-visual-relationship-and-attributes-detection}, scene graph generation \cite{scene-graph-generation-from-objects-phrases-and-region-captions, graph-rcnn-for-scene-graph-generation, neural-motifs, vip-net, factorizable-net, zoom-net}, etc. 

In this paper, we exclude visually-irrelevant relationships in VG and construct new Visually-Relevant Relationships dataset (VrR-VG). Rather than suffering from visually irrelevant relationships and easily predictable without visual information, VrR-VG focus on the visually relevant relationships and offers more cognitive abilities for image representation.
%The previous datasets contain many visual-irrelevant relationships, which are easily predictable without visual information. In contrast, VrR-VG focuses on the visually-relevant relationships and offers more semantic information. 

\textbf{Representation Learning:} Numerous deep learning methods have been proposed for representation learning with various knowledge~\cite{KG1, KG2, TKSN, infomax}. In image representation, these methods offer two aspects in image understanding: one is object category level, the other is instance level. GoogLNet \cite{googlenet}, ResNet \cite{resnet}, Inception \cite{inception}, ResNext \cite{resnext}, etc. trained on Imagenet \cite{imagenet} focus on object category classification. Since the supervision are object categories, the methods tend to give a holistic representation of images and figure out the features with the salient instance attention. Furthermore, as it is common that multiple instances exist in images, focusing on the salient instance is not enough to represent the scene. To explore multiple instances, detection task provides some effective tools. Jin et al.~\cite{aligning-where-to-see} apply selective search \cite{selective-search} to give salience region proposals. A similar idea also appears in RCNN \cite{rcnn}, in which the network generates many region proposals first and work out detection result for every instance. Faster-RCNN \cite{faster-rcnn} further improves the idea of region proposals and provide a faster and more elegant method to limited region proposals. Based on region proposals, Peter et al. \cite{bottom-up} proposed a bottom-up and top-down attention method to represent images. They utilize the locations, categories, and attributes of instances to learn the representation and get improvement in several cognitive tasks. In our work, we go deeper into multiple instances representation by adding inter-instance relationships. All instance locations, categories, attributes, together with relationships are jointly utilized in representation learning.

\vspace{-.1in}
\section{Visually-relevant Relationships Dataset}
To identify visually-irrelevant relationships, a hypothesis is proposed first that, \textit{if a relationship label in different triplets is predictable according to any information except visual information, the relationship is visual-irrelevant.} For distinguishing visually-relevant relationships, we introduce a novel %visually-relevant relationships discriminator (V$\text{R}^2$D).  V$\text{R}^2$D
visual discriminator network (VD-Net). VD-Net is a tiny network to predicate relation labels according to entities' classes and bounding boxes without images. The relation labels, which are not highly predictive by VD-Net, would be regarded as visually-relevant relationships. After reducing duplicate relationships by hierarchical clustering and filtering out the visually-irrelevant relationships, we constructed a new dataset named Visually-relevant Relationships Dataset (VrR-VG) from VG.

\begin{table*}[!t]
\centering
%\small
\resizebox{1.4\columnwidth}{!}
{
\begin{tabular}{|l|r|r|r|r|r|r|}
\hline
Dataset & \textit{object} & \textit{bbox}  & \textit{relationship} & \textit{triplet}  & \textit{image}  \\ \hline\hline
Visual Phrase \cite{VisualPhrases} & 8          & 3,271     & 9           & 1,796       & 2,769   \\
Scene Graph \cite{image-retrieval-using-scene-graphs}   & 266        & 69,009    & 68          & 109,535     & 5,000   \\
VRD \cite{visual-relationshp-detection-with-language-priors}           & 100        & -         & 70          & 37993       & 5,000   \\
Open Images \cite{open-image}   & 57         & 3,290,070 & 10          & 374,768     & -       \\
Visual Genome \cite{visual-genome} & 33,877     & 3,843,636 & 40,480      & 2,347,187   & 108,077 \\
VG150 \cite{iterative-message-passing}         & 150        & 738,945   & 50          & 413,269     & 87,670  \\
VrR-VG (ours)         & 1,600      & 282,460   & 117         & 203,375     & 58,983  \\ \hline
\end{tabular}
}
%\caption{Visual relationship datasets comparison. We compare the number of instance categories ($s/o$), single instance annotations ($s/o$ $instance$), relation predicates ($r$), unique relation triplets ($\left \langle  \textit{s, r, o} \right \rangle$), and images ($image$) in different datasets. %\textcolor{blue}{VrR-VG has competitive data scale and harder to predict without visual information as shown in \ref{sec:exp-data_cmp}.}
\caption{Visual relationship datasets comparison. We compare the number of object categories ($object$), single instance annotations ($bbox$), relationship categories ($relationship$), unique relation triplets ($triplet$), and images ($image$) in different datasets. %\textcolor{blue}{VrR-VG has competitive data scale and harder to predict without visual information as shown in \ref{sec:exp-data_cmp}.}
}
\label{dataset_table}
\vskip -1em
\end{table*}

%In this section, we introduce a novel visually-relevant relationship discriminator (V$\text{R}^2$D). V$\text{R}^2$D is a simple fully connected network aims to recognize relation labels by entities' classes and bounding boxes without images inputs. The relation labels, which are highly predictive by V$\text{R}^2$D, would be regarded as visually-irrelevant relationships. After reducing duplicate relationships by hierarchical clustering and filtering out the visually-irrelevant relationships, we constructed a new dataset named Visually-relevant Relationships Dataset (VrR-VG) from VG. 

%\subsection{Visually-relevant Relationship Discriminator}
\begin{figure}
    \centering
    \includegraphics[page=15,width=0.7\linewidth]{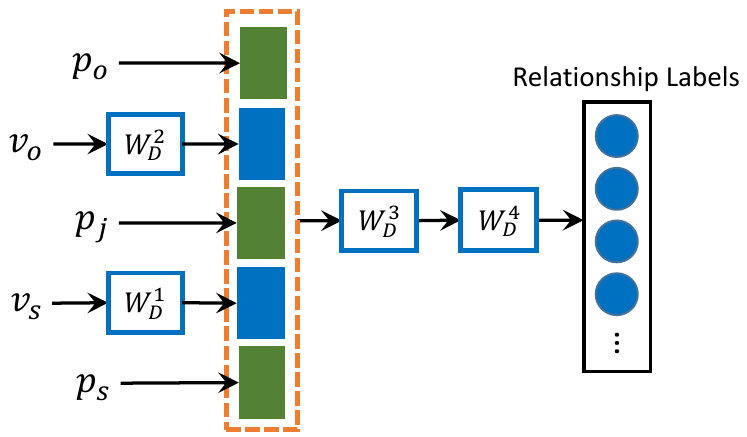}
    \caption{Structure of visual discriminator (VD-Net). With the information of instances' categories and locations, this tiny network is already able to predict most of the visually-irrelevant relationships with high accuracy.
    }
\label{trd_fig}
\end{figure}

\subsection{Visual Discriminator: VD-Net}
%To distinguish visually-irrelevant relationships, a hypothesis is proposed that, \textit{if a relation predicate is predictable according to any information except visual information, the relationship is visually-irrelevant}. 
In our work, a simple visual discriminator network (VD-Net) is proposed for selecting visually-irrelevant relationships. To prevent the overfitting, the network structure design follows the guideline of ``tinier is better''. Our VD-Net aims to recognize relationships without visual information from images. %Specifically, the inputs are word vectors of instances' categories learned from Natural Language corpus and the bounding boxes. GloVe \cite{glove} is applied for word vector. 

Each bounding box of instance in the image can be defined by a four-tuple $p=\{x, y, h, w\}$ that specifies its top-left
corner $(x, y)$, height $h$ and width $w$. The position embedding of object and subject can be represented as four-tuple $p_o$ and $p_s$ respectively, where $p_o = \{ x_{o}, y_{o}, h_{o}, w_{o}\}$ and $p_s = \{x_{s}, y_{s}, h_{s}, w_{s}\}$. The bounding boxes of given object and subject in related entities are embedded to a jointly vector as following equation:

\begin{equation}
\begin{aligned}
    p_j &= [ o_{x}, o_{y}, w_{o}, w_{s}, h_{o}, h_{s}, \frac{c_{s}-c_{o}}{w_{s}}, \frac{c_{s}-c_{o}}{h_{s}},\\
    & ( \frac{c_{s}-c_{o}}{w_{s}})^2, ( \frac{c_{s}-c_{o}}{h_{s}} )^2, \log(\frac{w_{o}}{w_{s}} ),
    \log( \frac{h_{o}}{h_{s}}) ]
\end{aligned}
\end{equation}
where $o_{x}, o_{y}$ are offsets of boxes computed by the difference between the coordinates of subject and object, $[w_o, h_o]$ and $[w_s, h_s]$ are width and height of bounding boxes, and $[c_{o}^x, c_{o}^y]$ and $ [c_{s}^x, c_{s}^y]$ are the center coordinates of the boxes.
%Above position embeddings provide position information for the network. 

The details of VD-Net are given in Fig.~\ref{trd_fig} where $v_s$ and $v_o$ are the word vectors of subject and object categories. GloVe \cite{glove} is applied for initializing word embeddings. $W_D^*$ are learnable weights. After a fully-connected layer, instance categories' features are concatenated with position embedding $p_o, p_s$ and $p_j$ correspondingly. %The word vectors are processed by a simple fully-connected layer. Then, the output features are concatenated with position embedding $p_o, p_s$ and $p_j$. %To fuse all the information, the subject and object features are concatenated with the jointly positional embedding. 
Finally, another two fully-connected layers and batch normalization layers are applied for classifying relation labels. %, whose input dimensions are 502 and 400. 
%The batch normalization layer also applied after the last fully-connected layer. 
We discard relationships which have larger accuracy than a threshold $\alpha$, and those reserved relationships are selected for generating the dataset. In this paper, we set $\alpha$ as 50\% due to the trade-off between dataset scale and visually-relevant quality.

The VD-Net merely contains three fully-connected layers, but it is already sufficient to predict most of the visually-irrelevant relationships, like ``wear", ``on", ``above", etc. 
%It is worth noting that 
More than 37\% of relation labels in VG150 can be predicted with at least 50\% accuracy by using such a crude neural network without any visual information.

\subsection{Dataset Construction}\label{dataset_construction}
We pre-process VG and extract top 1600 objects and 500 relationships to generate a basic data split. The raw relation labels in VG contain many duplications, such as ``wears" and ``is wearing a", ``next" and ``next to". Those labels may confuse the network because all those labels are correct to the same object and subject combination. We represent the labels by GloVe word vector, and filter out the duplicate relationships by applying hierarchical clustering~\cite{hierarchical_cluster} on relationships' word vectors. This simple operation reduces label categories from 500 to 180. We named this dataset after clustering as \textbf{R-VG}. Then, to exclude visually-irrelevant relationships, the VD-Net is utilized to train and evaluate with the 180 relationship labels in R-VG. Finally, we get 117 relation labels as \textbf{VrR-VG} relationships. It means our constructed VrR-VG is the subset of R-VG but filtered out the visually irrelevant relationships.

%\section{Informative Visual Representation Learning}
\section{Relationship-Aware Representation Learning}

\begin{figure}
    \centering
    \includegraphics[page=9,width=0.85\linewidth]{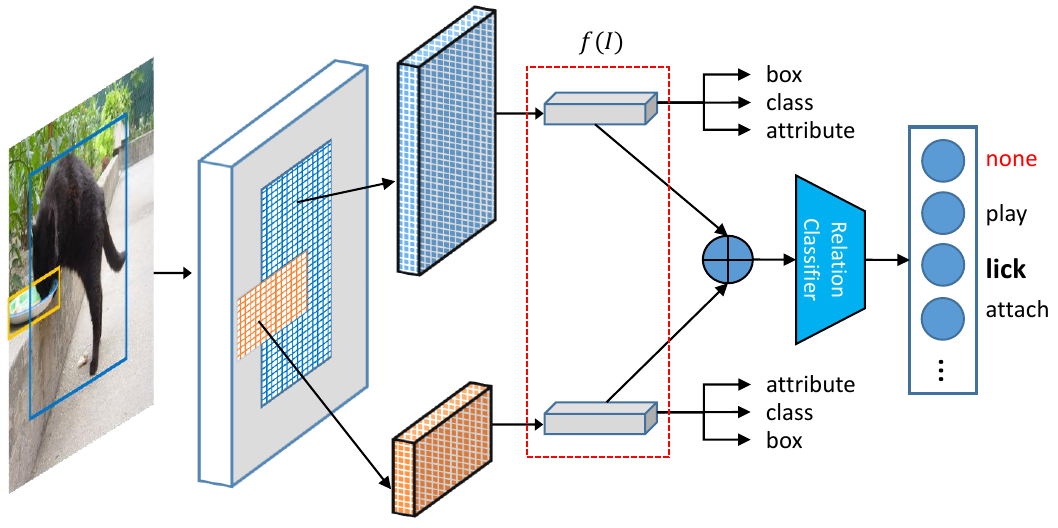}
    \caption{Overview of our proposed relationships-aware representation learning method. The feature vectors in the red box are learned visual representations for instances in the image. All the single instance properties and relationships among instances are utilized and embedded into features, which energizes features more cognitive abilities.}
\label{feat_fig}
\vskip -1em
\end{figure}

As shown in Fig.~\ref{feat_fig}, to model entire visual information in an image, the properties of isolated instances like category, position, attribute and the interaction of related instances are all useful. In our framework, all the properties are utilized for training features. We extract single instances proposals, and then train the model with all properties in images.
%supervision of the categories, locations, attributes, and relationships.

In detail, for the detector setting for single instances, Faster-RCNN~\cite{faster-rcnn} with ResNet101~\cite{resnet} is used as instance detector in our framework. We apply Non-maximum suppression (NMS) operation on regions proposals and then select $k$ candidate proposals according to IOU threshold. Then, through a mean-pooling layer, proposals' features $f(I)$ are integrated into the same dimensions.%, where $i=1, 2, ..., k$.

To learn the single instance properties, together with original detection operation, we set a classifier to learn instance attributes. The overall isolated properties are learned as follow: 
\begin{equation}
\begin{aligned}[b]
LOC_i =& W_{loc}^T f(I)+ b_{loc},  \\
CLS_i =& W_{cls}^T f(I) + b_{cls},  \\
ATT_i = W_{attr2}^T (W_{attr1}^T &[CLS_i, f(I)] + b_{attr1}) + b_{attr2}
\end{aligned}
\end{equation}
where $W_{loc}$, $W_{cls}$, $W_{attr1,2}$, $b_{loc}$, $b_{cls}$ and $b_{attr1,2}$ are learnable parameters, $[*]$ is concatenate operation. $LOC_i$, $CLS_i$, and $ATT_i$  are the bounding boxes, classes and attribute predictions for the $i$-th instance. We learn the relation representation by the following equation:

\begin{equation}
\begin{aligned}
& N_i = W_{R1} f(I) + b_{R1}, \\
& R_{i,j} = W_{R2}(N_i + N_j) + b_{R2}
\end{aligned}
\end{equation}
where $W_{R*}$ and $b_{R*}$ are learnable parameters for mapping instance to relation domain, $N_i$ is the node after mapping, and $R_{i,j}$ is the relation prediction between the proposal instances $i$ and $j$.

Formally, in training procedure, locations, categories, attributes of single entities and the relationships participate and supervise visual representation learning. The proposal features of single instances are extracted from the detector first. Then, the features are mapped into the relationship space. We fuse the mapped features to get relation predictions between proposals. Since there are $k$ proposals in our works, all the $k\times (k-1)$ combinations participate in features training. As a result, the feature contains all the information of isolated instances and the interaction among instances. We utilize the final features on VQA and image captioning tasks and evaluate the performance gains.

\section{Experiments}
% Please add the following required packages to your document preamble:
% \usepackage{multirow}

In this section, we discuss the properties of our data split from two aspects. One is the datasets comparison, the other is dataset quality evaluation by applying the visual representations learned from different datasets on cognitive tasks like VQA and image captioning. 

\subsection{Datasets Comparison}
\label{sec:exp-data_cmp}

\subsubsection{Relationships Analysis}
%To estimate the relationships reserved in our split, we compare VrR-VG dataset with other relation datasets on scene graph generation task. As we known, 

\begin{figure}[t!]
    \centering
    \small 
    \includegraphics[page=12,width=0.8\linewidth]{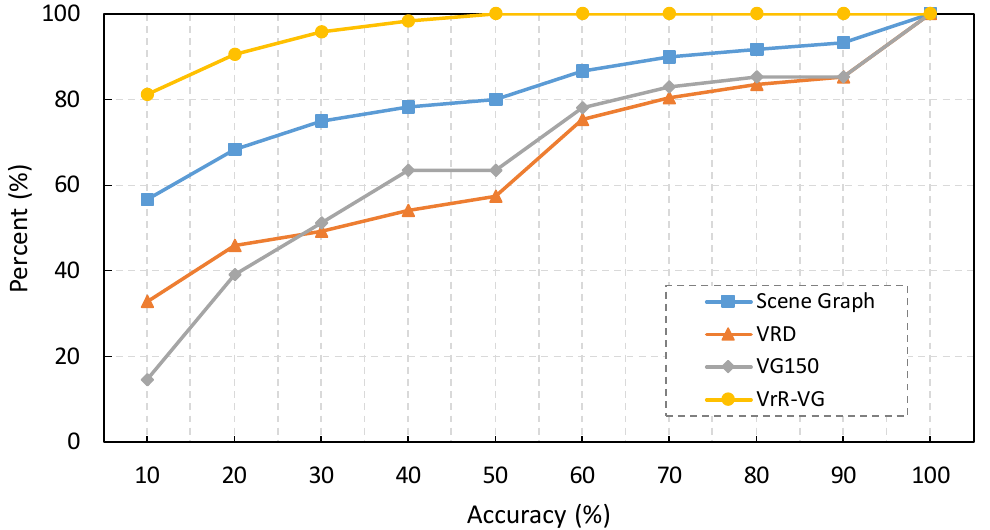}
    \caption{Accuracy proportions in different datasets by the VD-Net. The vertical axis indicates the accumulative proportions. Different from previous relation datasets, most relationship labels in our VrR-VG are unpredictable without image inputs.}
    \label{visual-relevant-cmp}
\vskip -1em
\end{figure}

%In the comparison of image amount, the VG150 data split, which was adapted in many relationships representation works \cite{neural-motifs, iterative-message-passing},
%contains 87670 images and 588586 triplet pairs. VrR-VG has 58983 images and 23375 relation pairs, which is less than VG150.
%contains 87670 images, 588586 different kinds of relation triplets in total. While VrR-VG has 58983 images with 23375 unique triplets and 203375 triplets in total, which are less than VG150. 

We compare the accuracy distributions of relationships predicted by VD-Nets trained on different scene graph datasets in Fig.~\ref{visual-relevant-cmp}. We can find that 75\%, 20\%, 42\% and 37\% of relationships in Visual Phrase dataset, Scene Graph dataset, VRD dataset, and VG150 have more than 50\% accuracy in relation predicates prediction with VD-Net respectively, which only depends on instances' locations and categories. Apparently, VrR-VG is more visually-relevant than others. It also means that VrR-VG is far harder than others in predicting relation predicates without visual information from images. 

\begin{figure}[htbp!]
    \centering
    \includegraphics[page=1, width=\linewidth]{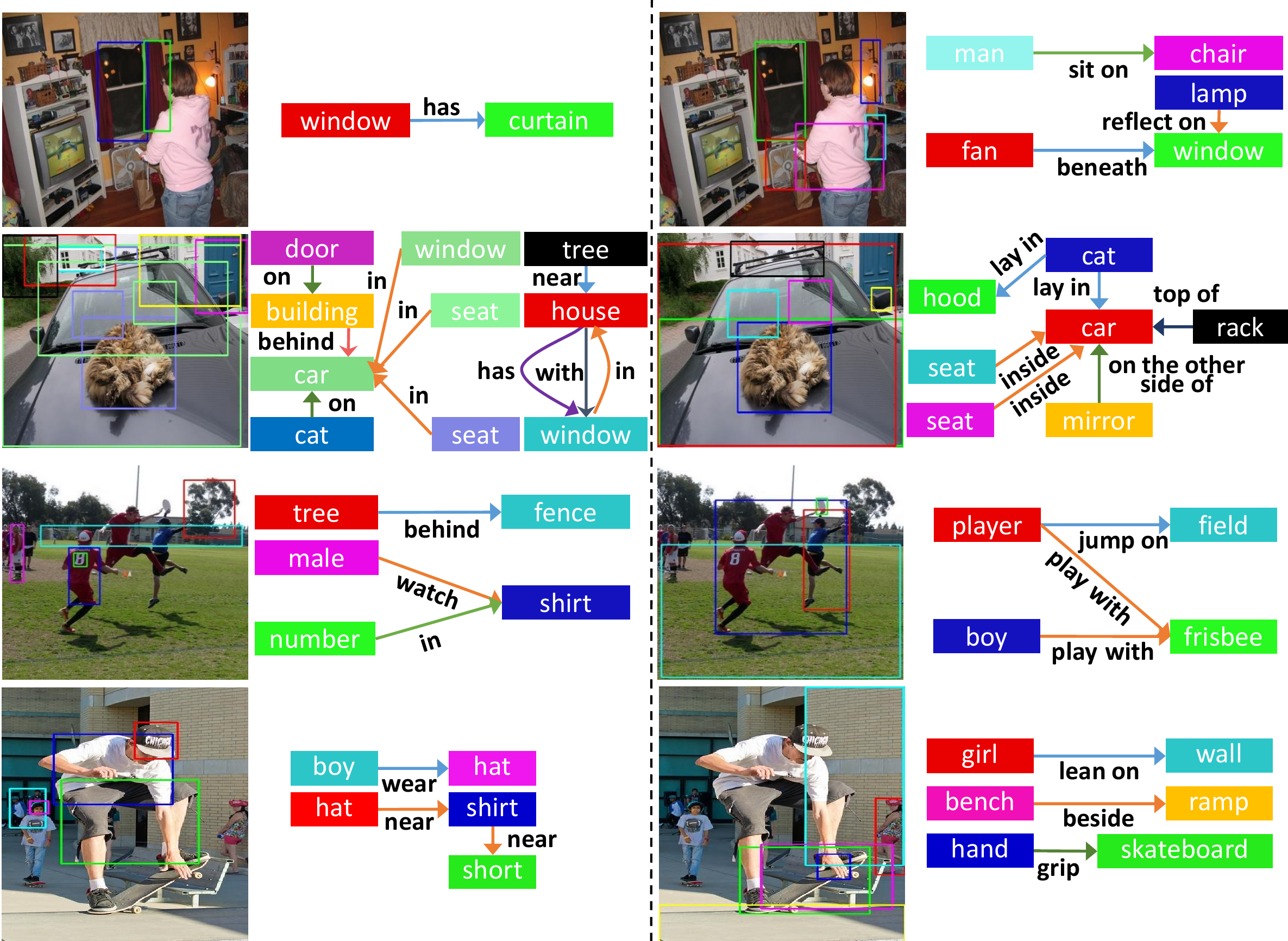}
    \caption{Examples of scene graph in VG150 (left) and VrR-VG (right). More diverse and visually-relevant relationships are contained in VrR-VG. }
    \label{additional_sg_exp}
\vskip -1em
\end{figure}

%Moreover, as shown in Figure~\ref{dataset_rela_distribution}, the distribution of our dataset is more balance. Since VG150 is generated depending on the labels' frequency, the easy and general relationships are inevitably involved. labels like ``on", ``of", etc. in VG150 are too trivial to describe entities interaction. 
As shown in Fig.~\ref{dataset_rela_distribution}, 
top-12 relationship labels take 91.55\% of VG150 dataset. Meanwhile, most of these labels are spatial relationships which can be estimated merely by instances' locations.
%In the top-12 labels in VG150, most of labels are spatial relationships which can be estimate merely by instances' locations, meanwhile top-12 labels take 91.55\% of the overall dataset. %Some statistically biased relationships like "wears" and "wearing" are also easy to be predicted as shown in Figure~\ref{object_bias_eg}. 
%and there are also some relationships having the same semantic meaning appear in the top relationships in VG150, e.g. ``wears" and ``wearing" are regard as two different relationships and accounting for 11.87\% and 0.84\% in VG150 respectively.
Comparatively, our top-12 labels take 67.62\% and are more significant in the cognitive domain. Relationships like ``hanging on", ``playing with", etc. are hard to be estimated without enough understanding in corresponding scenes. Moreover, VrR-VG consist of 117 relationships is more diverse than the former 50 relationships in VG150. More scene graph examples from our VrR-VG are given in Fig.~\ref{additional_sg_exp}

%So, our data split is much more difficult in semantic representation in visual. We show an exemplary scene graph with in Figure~\ref{scene_graph_eg_intro}. It can be found that compares with previous VG150 dataset, the relationships in our dataset are more diverse and contain more semantic information for describing the scene. 

\subsubsection{Scene Graph Generation}

\begin{table*}[!t]
    \begin{center}
%\small
\resizebox{1.5\columnwidth}{!}
{
\begin{tabular}{|c|c|c|c|c|c|c|c|c|}
\hline
\centering
\multirow{3}{*}{Methods}         & \multicolumn{8}{c|}{Datasets}                                              \\ \cline{2-9} 
                                 & \multicolumn{4}{c|}{Method specific VG splits} & \multicolumn{4}{c|}{VrR-VG}       \\ \cline{2-9} 
                                 & Metrics   & SGDet  & SGCls  & PredCls  & Metrics & SGDet & SGCls & PredCls \\ \hline
\multirow{2}{*}{MSDN~\cite{scene-graph-generation-from-objects-phrases-and-region-captions}}            & R@50       & 11.7   & 20.9   & 42.3     & R@50     & 3.59  & -     & -       \\ %\cline{2-9} 
                                 & R@100      & 14.0   & 24.0   & 48.2     & R@100    & 4.36  & -     & -       \\ \hline %\cline{2-9} 
                                 %& R-gap     & 2.3    & 3.1    & 5.9      & R-gap   & 0.77  & -     & -       \\ \hline
\multirow{2}{*}{Vtrans~\cite{visual-translation-embedding-network-for-visual-relation-detection}}          & R@50       & 5.52   & -      & 61.2     & R@50     & 0.83 & -     & 44.69   \\ %\cline{2-9} 
                                 & R@100      & 6.04   & -      & 61.4     & R@100    & 1.08   & -     & 44.84   \\ \hline %\cline{2-9} 
                                 %& R-gap     & 0.52   & -      & 0.26     & R-gap    & 0.25   & -     & 0.15    \\ \hline \hline
%\multirow{3}{*}{methods}          & \multicolumn{8}{c|}{datasets}                                              \\ \cline{2-9} 
\multirow{2}{*}{Methods} & \multicolumn{4}{c|}{VG150}             & \multicolumn{4}{c|}{VrR-VG}       \\ \cline{2-9} 
                                 & Metrics   & SGDet  & SGCls  & PredCls  & Metrics & SGDet & SGCls & PredCls \\ \hline
\multirow{2}{*}{Neural-Motifs~\cite{neural-motifs}}   & R@50       & 27.2   & 35.8   & 65.2     & R@50     & 14.8  & 16.5  & 46.7    \\ %\cline{2-9} 
                                 & R@100      & 30.3   & 36.5   & 67.1     & R@100    & 17.4  & 19.2  & 52.5    \\ \hline %\cline{2-9} 
                                 %& R-gap     & 3.1    & 0.7    & 1.9      & R-gap   & 2.6   & 2.7   & 5.8     \\ \hline
\multirow{2}{*}{Message Passing~\cite{iterative-message-passing}} & R@50       & 20.7   & 34.6   & 59.3     & R@50     & 8.46  & 12.1  & 29.7    \\ %\cline{2-9} 
                                 & R@100      & 24.5   & 35.4   & 61.3     & R@100    & 9.78  & 13.7  & 34.3    \\ \hline %\cline{2-9} 
                                 %& R-gap     & 3.8    & 0.8    & 2.0      & R-gap   & 1.3   & 1.6   & 4.6     \\ \hline
\end{tabular}
}
\end{center}
\caption[Caption for LOF]{The performance of different methods for scene graph generation on different datasets.
%R-gap (recall gap) indicates the difference between R@100 and R@50. 
The MSDN and Vtrans methods are evaluated in the other data splits, which are also split from VG by frequency. While Neural-Motifs and Message Passing methods use the same VG150 data split. Additionally, evaluating details about SGCls and PredCls in MSDN and SGCls in Vtrans are not released, so some numbers are not reported in our experiments. 
%All the performances of methods decrease in our dataset, \textcolor{blue}{which indicates that the previous datasets with visually-irrelevant relationships are easier to predict than VrR-VG.}
}
\label{sg_table}
\end{table*}

Since scene graph generation task points to the representability of relationships directly, we also evaluate and compare the task performances in VrR-VG with others datasets by using different widely used scene graph generation methods, including MSDN~\cite{scene-graph-generation-from-objects-phrases-and-region-captions}, Vtrans~\cite{visual-translation-embedding-network-for-visual-relation-detection}, Message Passing~\cite{iterative-message-passing} and Neural-Motifs~\cite{neural-motifs}. We evaluate following metrics~\cite{visual-relationshp-detection-with-language-priors, neural-motifs} with R@50 and R@100\footnote{R@$N$: the fraction of times the correct relationship is predicted in the top-$N$ predictions.} in scene graph generation: 

\begin{itemize}
%\vspace{-1.2pt}
\setlength{\itemsep}{0pt}
\setlength{\parskip}{0pt}
\setlength{\parsep}{0pt}
\item Scene Graph Detection (SGDet): given images as inputs, predict instance locations, categories, and relationships.
\item Scene Graph Classification (SGCls): given images and instances locations, predict instance categories and relationships.
\item Predicate Classification (PredCls): given images, instance locations, and categories, predict relationships.
\item Predicate detection (PredDet): given images, instance locations, categories, and relationship connections, predict relationship labels. 
\end{itemize}
 %VG150~\cite{neural-motifs, iterative-message-passing}. Additionally, Neural-Motifs~\cite{neural-motifs}, Message Passing~\cite{iterative-message-passing} use pre-trained Faster-RCNN with VGG backbone as object detector. The mAP of the detector is 20.0 and 8.2 in VG150 and our VrR-VG at 50\% IOU. 
\vspace{-4.8pt}
%With the experiments in some relation representation methods, 
As shown in Table~\ref{sg_table} %and~\ref{predicate_det_label}
, the performances apparently decrease when using our dataset. With the relationships selected by our method, the scene graph generation task becomes more difficult and challenging.%, \textcolor{blue}{which confirms that the previous datasets can be better solved with many visually-irrelevant relationships involved. }

\begin{table}[!t]
    \begin{center}
%\small
\resizebox{0.9\columnwidth}{!}
{
\begin{tabular}{|c|c|c|c|c|}
\hline
Methods                             & Metrics & VG150 & VrR-VG & $\Delta$ \\ \hline \hline
\multirow{2}{*}{Message Passing}    & R@50     & 93.5  & 84.9  & \textbf{8.6} \\ %\cline{2-4} 
                                    & R@100    & 97.2  & 91.6  & \textbf{5.6} \\ \hline
\multirow{2}{*}{Frequency-Baseline} & R@50     & 94.6  & 69.8  & \textbf{24.8} \\ %\cline{2-4} 
                                    & R@100    & 96.9  & 78.1  & \textbf{18.8} \\ \hline
\multirow{2}{*}{Neural-Motifs}      & R@50     & 96.0  & 87.6  & \textbf{8.4} \\ %\cline{2-4} 
                                    & R@100    & 98.4  & 93.4  & \textbf{5.0} \\ \hline
\end{tabular}
}
\end{center}
\caption[Caption for LOF]{Evaluation results of different datasets in PredDet. $\Delta$  indicates the performance gap between different datasets. The results show that the relation representation problem in our dataset is solvable and the learnable methods apparently do better than statistical method. Meanwhile, the high requirement is put forward in our dataset}
\label{predicate_det_label}
\end{table}

\begin{table*}[!t]
%\small 
\begin{center}\resizebox{1.8\columnwidth}{!}
{
\begin{tabular}{|l|l|l|l|l|l|}
\hline
Dataset & Object Category & Object Annotation & Attribute Category & Attributes Annotation &  Image   \\ \hline\hline
BottomUp-VG~\cite{bottom-up} & 1600              & 3,404,999          & 400                  & 1,829,438              & 107,120 \\ \hline
VrR-VG  & 1600              & 2,106,390          & 400                  & 1,109,650              & 58,983  \\ \hline
\end{tabular}
}
\end{center}
\caption{The detail statistics of BottomUp-VG and VrR-VG. }
\label{bottomupvg_vrrvg}
\end{table*}

\begin{table*}[!ht]
    \begin{center}
%\small
\resizebox{1.8\columnwidth}{!}
{
\begin{tabular}{|c|c|c|c|c|c|c|c|}
\hline
VQA Method & Feature Learning Method & Used Relation & Dataset   &  Yes/No & Numb. & Others & All   \\ \hline \hline
%\multirow{2}{*}{Bottom-Up~\cite{bottom-up}} & \ding{56} & Bottom-Up & 80.3  & 42.8  & 55.8  & 63.2 \\ %\hline
% & \ding{56} & ImageNet & 77.6  & 37.7  & 51.5  & 59.4 \\ \hline
\multirow{5}{*}{\begin{tabular}[c]{@{}c@{}}MUTAN~\cite{mutan}\end{tabular}} & \multirow{2}{*}{\begin{tabular}[c]{@{}c@{}}BottomUp~\cite{bottom-up}\end{tabular}} & \ding{56}           & BottomUp-VG & 81.90  & 42.25 & 54.41 & 62.84 \\ \cline{3-8}  %\hline
& & \ding{56}           & VrR-V$\text{G}_{obj}$  & 80.46  & 42.93 & 54.89 & 62.93 \\ \cline{2-8} %\hline
& \multirow{3}{*}{\begin{tabular}[c]{@{}c@{}}Ours\end{tabular}} & \textcolor{red}{\ding{52}}          & VG150   & 79.00  & 39.78 & 49.87 & 59.49 \\ \cline{3-8} %\hline
& & \textcolor{red}{\ding{52}}         & R-VG   & 82.35  & 43.91 & 54.89 & 63.77 \\  \cline{3-8}
& & \textcolor{red}{\ding{52}}         & VrR-VG   & \textbf{83.09}  & \textbf{44.83} & \textbf{55.71} & \textbf{64.57} \\ \hline
\multirow{5}{*}{\begin{tabular}[c]{@{}c@{}}MFH~\cite{mfh}\end{tabular}} & \multirow{2}{*}{\begin{tabular}[c]{@{}c@{}}BottomUp\end{tabular}} & \ding{56}           & BottomUp-VG  & 82.47  & 45.07 & 56.77 & 64.89 \\ \cline{3-8} %\hline
& & \ding{56}& VrR-V$\text{G}_{obj}$     & 82.37  & 45.17 & 56.40 & 64.68 \\ \cline{2-8}%\hline
& \multirow{3}{*}{\begin{tabular}[c]{@{}c@{}}Ours\end{tabular}} & \textcolor{red}{\ding{52}}          & VG150     & 78.86  & 38.32 & 50.98 & 59.80 \\  \cline{3-8}%\hline
& & \textcolor{red}{\ding{52}}         & R-VG    & 82.43  & 43.70 & 55.81  & 64.22 \\\cline{3-8}
& & \textcolor{red}{\ding{52}}         & VrR-VG    & \textbf{82.95}  & \textbf{45.90} & \textbf{57.34} & \textbf{65.46} \\ \hline
\end{tabular}
}
\end{center}
\caption{Comparison of features trained from different datasets for open-ended VQA on the validation split of VQA-2.0.
%Comparison between different methods based on image features learned from different datasets for open-ended VQA on the validation split of VQA-2.0 dataset. 
Features learned from our VrR-VG outperform all other relation datasets.}
\label{vqa_res}
\vskip -1em
\end{table*}

Notably, as the metric excluding the influence of detector performances, the relation predicates detection use paired detection ground truth for inputs and show the theoretical optimal performance in scene graph generation. As experimental results in Table~\ref{predicate_det_label}, the gaps of performances between statistical and learnable methods are notably larger. The values of R@50 and R@100 in Frequency-Baseline are merely 69.8 and 78.1, which are far from results in VG150. This means the frequency-based method does not work anymore in VrR-VG. Experiments reflect the previously proposed methods really ``learn'' in VrR-VG, instead of using visually-irrelevant information to fit the data defeats.

%Notably, \textcolor{blue}{as the metric excluding the influence of detector performances, the relation predicates prediction use paired detection ground truth for inputs and show the theoretical optimal performance in scene graph generation. As experimental results in Table~\ref{predicate_det_label}, the gaps of performances between statistical and learnable methods are notably larger. VrR-VG has 3-4 times larger gaps in learnable or unlearnable methods than VG150. The differences show that relationships are truly and better learned in VrR-VG, rather than statistically predicting by data bias.}
%\textcolor{blue}{Moreover, as shown in Table \ref{predicate_det_label}, although the results of Message Passing and Neural-Motifs decrease in our dataset, they also get convincible result. The facts indicate that the relationships selected in our dataset are representable and with fine-designed methods, problems in our split are not incompatible and infeasible. In contrary, the values of R@50 and R@100 in Frequency-Baseline are merely 69.8 and 78.1, which are far from results in VG150. This means the frequency-based method does not work anymore in VrR-VG. Experiments reflect the methods are really ``learn'' in VrR-VG, instead of using visually-irrelevant information to fit the data defeats.}

\subsection{Relationship-Aware Representation on Cognitive Tasks}
To evaluate the relation quality in cognitive level, we choose VQA and image captioning in experiments and apply the visual features learned from our constructed dataset on these cognitive tasks. We also compared our relationship-aware representation learning method with the previous instance level representation learning method Bottom-Up~\cite{bottom-up}. We named the dataset used in Bottom-Up as BottomUp-VG, which is also collected from VG dataset. The detail statistics of BottomUp-VG and VrR-VG are shown in Table~\ref{bottomupvg_vrrvg}. The experimental results of feature learned by Bottom-Up and our relationship-aware representation learning method are shown as ``Not Used Relation'' and ``Used Relation'' in Table~\ref{vqa_res} and Table~\ref{caption_table} respectively. To be fair, our proposed relationship-aware representation learning method follows the basic settings in Bottom-Up~\cite{bottom-up}. The experimental results demonstrate that the visually-relevant relationship plays an important role in high-level visual understanding.

%For the details,
%of our proposed informative visual representation method, 
%we follow the settings of bottom-up and top-down method~\cite{bottom-up}
%, we set $k=32$ for all experiments in this section and image features are integrated into 2048 dimensions by mean-pooling. %We train different features in original bottom-up and top-down method, our proposed method 
Additionally, we introduce a variant dataset VrR-V$\text{G}_{obj}$, which is based on VrR-VG but excludes relation data for ablation study. We apply our proposed feature learning for VrR-V$\text{G}_{obj}$ too, but without the weight of the relationship and relation loss is set as 0. %Another dataset R-VG contains all the visually-relevant and visually-irrelevant relationships from the original data split (all 180 relation labels before VD-Net as we descirbed in Section~\ref{dataset_construction}). The results from the features learned with the full-scale relationships are also presented in following sections. 

\begin{figure}[htbp!]
    \centering
    \includegraphics[page=1, width=\linewidth]{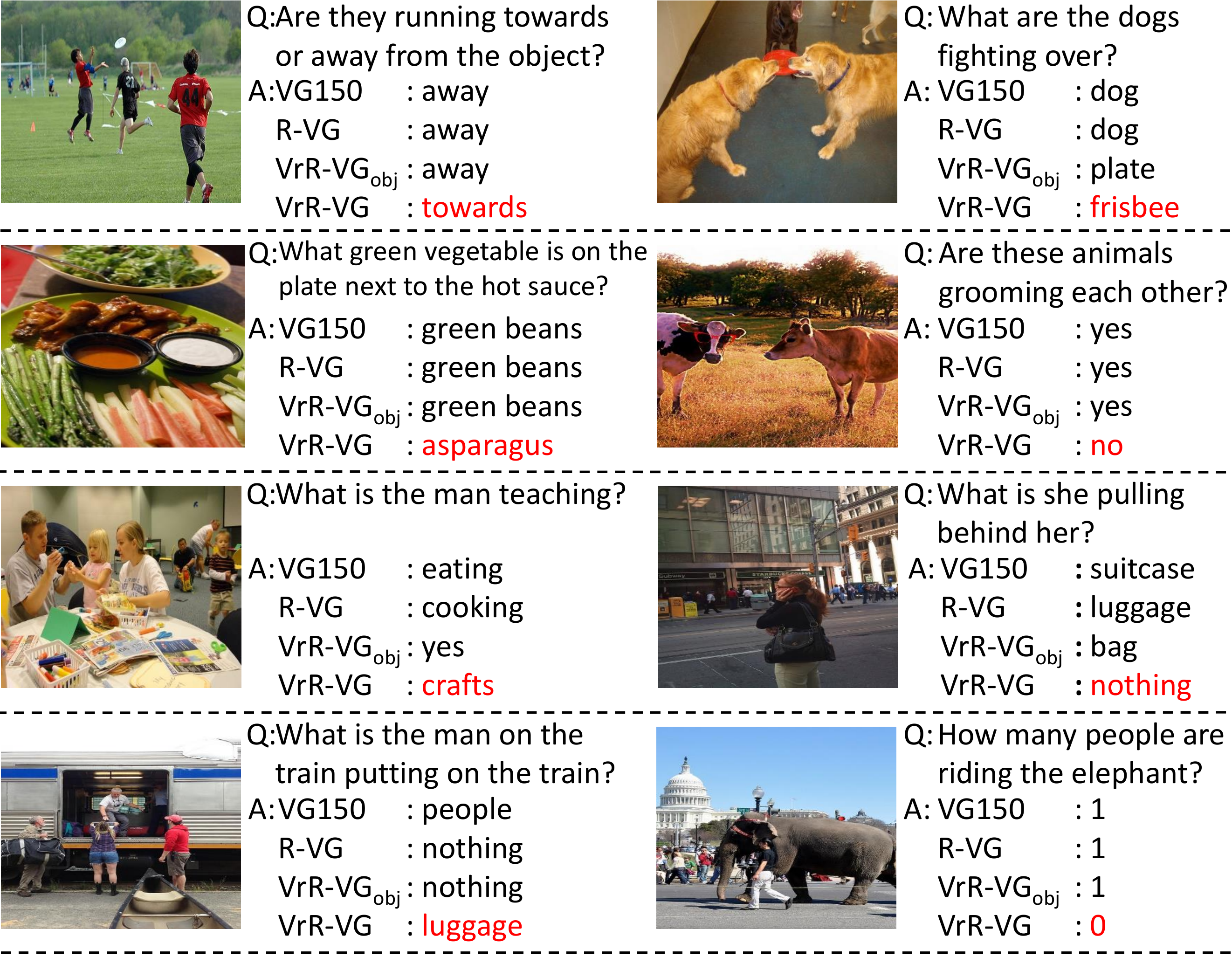}
    \caption{Examples of VQA. Features trained with VrR-VG provide more information for the interactions of instances. Best viewed in color.}
    \label{vqa_case}
\vskip -1em
\end{figure}

\vspace{-.1in}
\paragraph{VQA:} We applied two widely used VQA methods MUTAN~\cite{mutan} and MFH~\cite{mfh} for evaluating the quality of image feature learned from different datasets. Table~\ref{vqa_res} reports the experimental results on validation set of VQA-2.0 dataset~\cite{vqa2_dataset}. We can find that features trained with our VrR-VG obtain the best performance in all the datasets. We also compared the dataset used in Bottom-Up attention~\cite{bottom-up}, which is regarded as the strongest feature representation learning method for VQA.

With relation data, our VrR-VG performs better than dataset used in Bottom-Up attention and VrR-V$\text{G}_{obj}$. The results indicate that the relationship data is useful in VQA task, especially in the cognitive related questions as shown in Fig.~\ref{vqa_case}. It also demonstrates that our proposed informative visual representation method can extract more useful features from images. %Rather than yes/no or number categories, the main improvement appears in questions of others category. Since the others questions required more information from the interactions between objects, it is convincible and reasonable that our dataset achieves better performances. 
Besides, we also apply our proposed feature learning method on VG150 dataset. Since VG150 contains a majority of visually-irrelevant relationships which can be inferred easily by data bias as we mentioned, the features learned from VG150 usually lack the ability to represent complex visual semantics. 

Moreover, the experimental results also show that VrR-VG has better performance than R-VG, which demonstrates that filtering out visually-irrelevant relationship is beneficial to learning high-quality representations, and further demonstrate the merits of VD-Net.
%The higher quality of relation data energizes the features learned from our dataset and lead to a better performance in the open-ended VQA task.

\vspace{-.1in}
\paragraph{Image Captioning:}

\begin{table*}[!t]
\begin{center}
\resizebox{2.1\columnwidth}{!}
{
\begin{tabular}{|c|c|c|c|c|c|c|c|c|c|}
\hline
%\multicolumn{2}{|c|}{Methods}                                                                         & 
\begin{tabular}[c]{@{}c@{}} Image Captioning\\ Method\end{tabular} &
\begin{tabular}[c]{@{}c@{}} Feature  Learning\\ Method\end{tabular} &
\begin{tabular}[c]{@{}c@{}} Feature  Learning\\ Dataset\end{tabular} &
%Captioning Method & Feature Learnig  
Used Relation & BLEU-1 & BLEU-4 & METEOR & ROUGLE-L & CIDEr & SPICE \\\hline\hline
%& & & & & & & & & \hline \hline
\multirow{5}{*}{\begin{tabular}[c]{@{}c@{}}Cross-Entropy\\ Loss\end{tabular}} %& SCST~\cite{scst}                  & -                     & \ding{56}     & -      & 30.0   & 25.9   & 53.4     & 99.4  & -     \\ \cline{2-10} 
                            %                                                  & LSTM-A~\cite{lstm-a}                & -                     & \ding{56}     & 75.4   & 35.2   & 26.9   & 55.8     & 108.8 & 20.0  \\ \cline{2-10} 
                                                                              &
                                                           \multirow{2}{*}{BottomUp~\cite{bottom-up}}                                & BottomUp-VG        & \ding{56}     & \textbf{76.9}   & \textbf{36.0}   & 26.9   & 56.2     & 111.8 & 20.2  \\ \cline{3-10} 
                                                &                       & VrR-V$\text{G}_{obj}$            & \ding{56}     & 76.2   & 35.4   & 26.8   & 55.7     & 110.3 & 19.9  \\ \cline{2-10} 
                                                                                  & \multirow{3}{*}{Ours} & VG150             & \textcolor{red}{\ding{52}}  & 74.2   & 32.7   & 25.3   & 53.9     & 102.1 & 18.5  \\ \cline{3-10} 
                                                            
                                                                              &                       & R-VG                  & \textcolor{red}{\ding{52}}  & 76.3   & 35.4   & 27.0   & 56.0     & 111.2 & 20.0  \\ \cline{3-10} 
                                                                              &                       & VrR-VG                & \textcolor{red}{\ding{52}}  & \textbf{76.9}   & \textbf{36.0}   & \textbf{27.2}   & \textbf{56.3}     & \textbf{114.0} & \textbf{20.4}  \\ \hline
\multirow{5}{*}{\begin{tabular}[c]{@{}c@{}}CIDEr\\ Optimization\end{tabular}} %& SCST                  & -                     & \ding{56}     & -      & 34.2   & 26.7   & 55.7     & 114.0 & -     \\ \cline{2-10} 
                                                       %                       & LSTM-A                & -                     & \ding{56}     & 78.6   & 35.5   & 27.3   & 56.8     & 118.3 & 20.8  \\ \cline{2-10} 
                                                                              & \multirow{2}{*}{BottomUp}      & BottomUp-VG              & \ding{56}     & \textbf{79.6}   & 36.0   & 27.6   & 56.7     & 118.2 & 21.2  \\ \cline{3-10} 
                                                                                &                       & VrR-V$\text{G}_{obj}$            & \ding{56}     & 78.8   & 35.8   & 27.3   & 56.4     & 116.8 & 21.0  \\ \cline{2-10}
                                                                              & \multirow{4}{*}{Ours} & VG150                 & \textcolor{red}{\ding{52}} & 76.7   & 32.7   & 25.8   & 54.3     & 108.0 & 19.6  \\ \cline{3-10} 
                                                                             
                                                                              &                       & R-VG                  & \textcolor{red}{\ding{52}}  & 79.1   & 35.8   & 27.5   & 56.5     & 118.8 & 21.2  \\ \cline{3-10} 
                                                                              &                       & VrR-VG                & \textcolor{red}{\ding{52}}  & 79.4   & \textbf{36.5}   & \textbf{27.7}   & \textbf{56.9}     & \textbf{120.7} & \textbf{21.6}  \\ \hline
\end{tabular}
}
\end{center}
\caption[Caption for LOF]{Comparison of different single model with feature trained from different datasets for image captioning.
%Single model image captioning results with features learned from different datasets.
We evaluate the performances in MSCOCO 2014 caption dataset~\cite{ms-coco}. %The performances of features trained with our dataset is better than others.
}
\label{caption_table}
\vskip -1em
\end{table*}

%We adapt our constructed dataset VrR-VG and our proposed entity, attributes, localizations and interactions jointly feature representation learning method to image captioning task. 
Similar to the experiment process used in VQA task, we first generate the image features based on VG150, VrR-V$\text{G}_{obj}$, R-VG and VrR-VG respectively. Then we apply the caption model~\cite{bottom-up} for these image features with the same settings.

%The experiments results are shown in Table ~\ref{caption_table}. %The original Bottom-Up \cite{bottom-up} features have 10.2 MAP in detection. Features trained with our split and VG150 have 8.4 and 10.8 detection MAP. 
As shown in Table~\ref{caption_table}, we report the performances in VrR-VG and VG150 in both the original optimizer for cross entropy loss and CIDEr optimizer for CIDEr score. Features generated from our data split works better than VG150. All metrics in captioning have better performance when using both of the optimizers. %Meanwhile, considering the difference of detection map, our results have a little gap compared with original Up-down features in BLEU.
Moreover, in the comparison of adding relation or not, our complete VrR-VG has better performance than the VrR-V$\text{G}_{obj}$ and R-VG.
%which only contains objects information in original VrR-VG. 
This indicates that visually-relevant relationships are useful for image captioning. Despite the dataset BottomUp-VG has much more object annotations, attributes annotations and images than VrR-VG as shown in Table~\ref{bottomupvg_vrrvg},
relationship-aware representation learned from VrR-VG can still achieve comparable or better results with \textit{object, attribute} based representations learned from BottomUp-VG, owing to the visually-relevant relation information.
%The improvement shows the efficiency of our dataset and also reveal the power of relationships in semantic tasks. 

\begin{figure}[htb!]
    \centering
    \includegraphics[page=2, width=\linewidth]{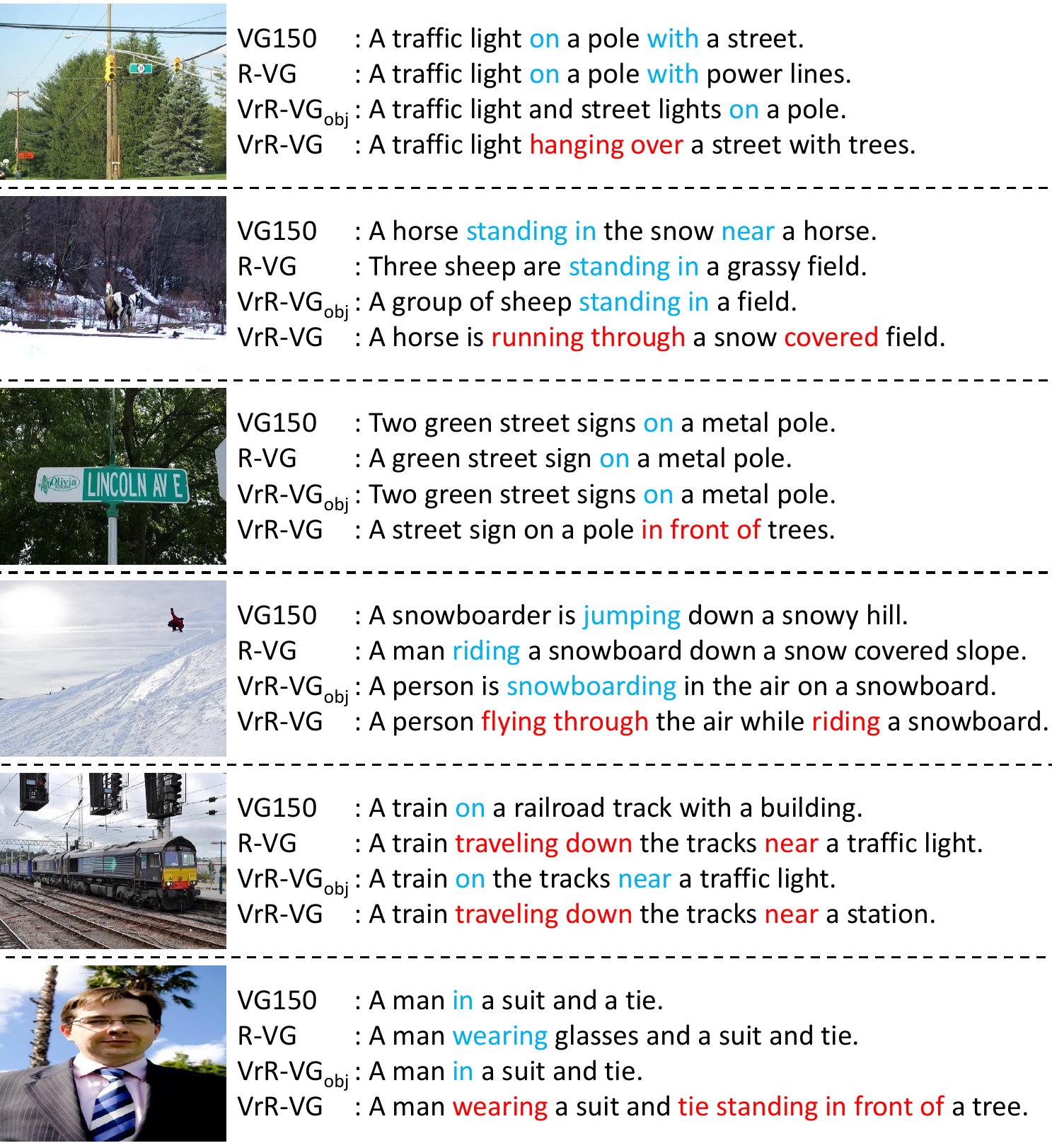}
    \caption{Examples of captioning. Features trained with relationships data offer more complex and diverse expression in predicates. Moreover, with visually-relevant relationships, more information about interactions among instances are also shown in results. Best viewed in color.}
    \label{caption_case}
\end{figure}

%We show some examples in Fig.~\ref{caption_case}, For fair comparison, all of these caption examples have accurate recognition in objects. %We focus on the difference of predicates, which can better reflect the influence of relation data. 
In examples of caption results as shown in Fig.~\ref{caption_case}, the features learned from our VrR-VG dataset lead to more diverse predicates and more vivid description than others. Rather than some simple predicates like ``on", ``with", etc., our features provide more semantic information and help models achieve more complex expression like ``hanging", ``covered", etc. Although \textit{this kinds of expressions may not lead to high scores in captioning metrics}, these vivid and specific results are valuable for cognitive tasks.

In total, the higher quality of relation data energizes the features learned from our dataset and leads to a better performance in the open-ended VQA and image captioning tasks.

\section{Conclusion}
A new dataset for visual relationships named Visually-relevant relationships dataset (VrR-VG) is constructed by filtering visually-irrelevant relationships from VG. 
Compared with previous datasets, VrR-VG contains more cognitive relationships, which are hard to be estimated merely by statistical bias or detection ground-truth. We also proposed an informative visual representation learning method learning image feature jointly considering entity labels, localizations, attributes, and interactions. The significant improvements in VQA and image captioning demonstrate that: (1) VrR-VG has much more visually-relevant relationships than previous relationship datasets, (2) visually-relevant relationship is helpful for high-level cognitive tasks, (3) our proposed informative visual representation learning method can effectively model different types of visual information jointly. 
\newline

\noindent\large\textbf{Acknowledgements:}
\normalsize
This work was in part supported by the NSFC (No.61602463, 61772407, 61732008, 61332018, u1531141), the National Key R\&D Program of China under Grant 2017YFF0107700, the World-Class Universities (Disciplines), the Characteristic Development Guidance Funds for the Central Universities No.PYl3A022, the Open Project Program of the National Laboratory of Pattern Recognition.

% Bibliography
%\bibliographystyle{ACM-Reference-Format}
%\bibliography{cite-file}

{\small
\bibliographystyle{ieee_fullname}
\bibliography{cite-file}
}

%\iffalse
\clearpage
\appendix
\onecolumn
\appendix
%\onecolumn
%\section{Supplementary Materials}
\section{Scene Graph Comparison}

\begin{figure*}[htbp!]
    \centering
    \includegraphics[page=1, width=\textwidth]{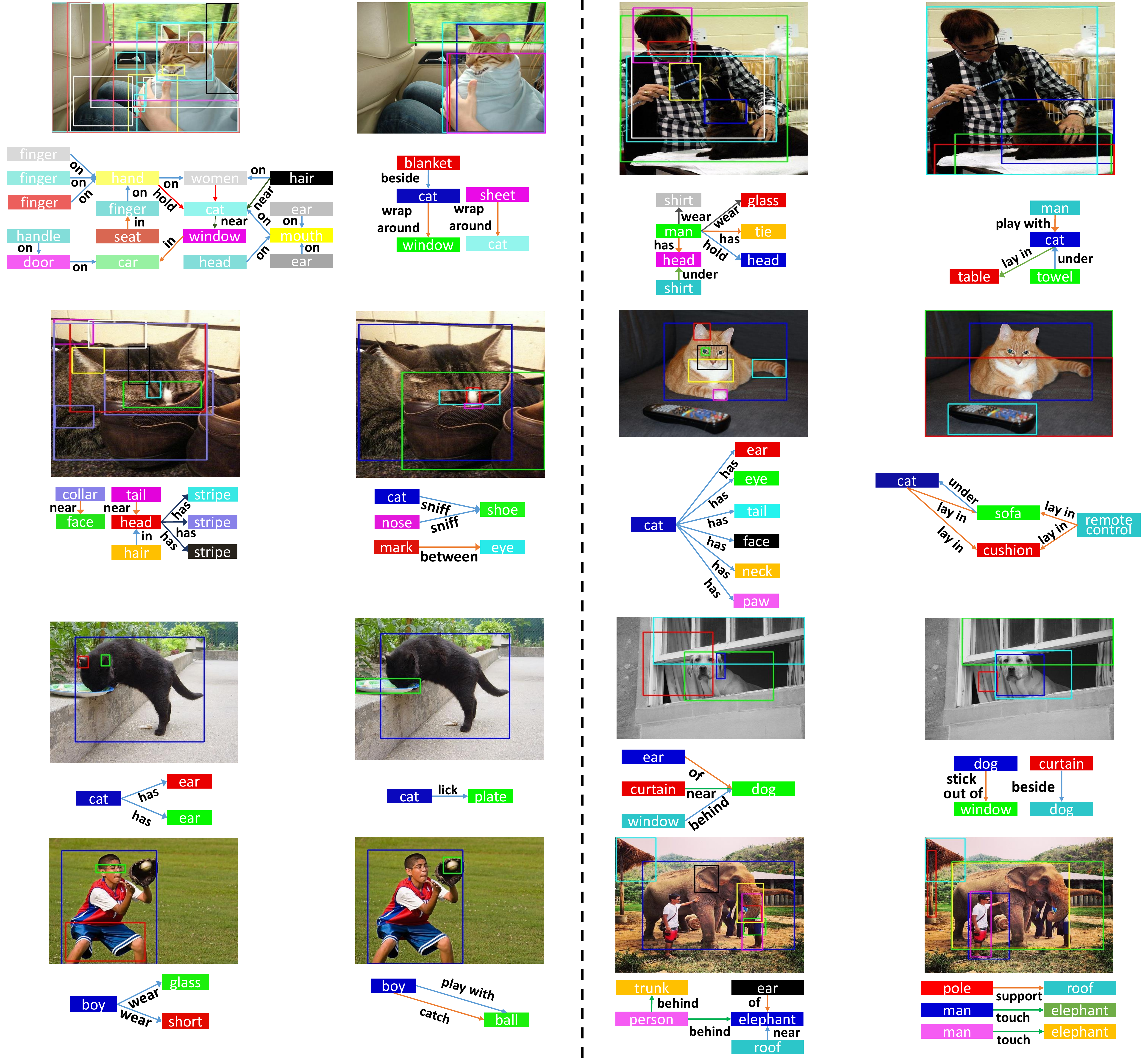}
    \caption{More examples of scene graph in VG150 (left) and VrR-VG (right).}
    \label{sg_eg_1}
\end{figure*}

We show additional scene graph examples from dataset VG150 and VrR-VG in Fig.~\ref{sg_eg_1}. Most of the visually-irrelevant relationships in VG150 like ``on'', ``near'', ``has'', etc. are excluded.

\FloatBarrier
\clearpage

\section{Results comparison on VQA task}

\begin{figure*}[htbp!]
    \centering
    \includegraphics[page=1, width=0.75\textheight]{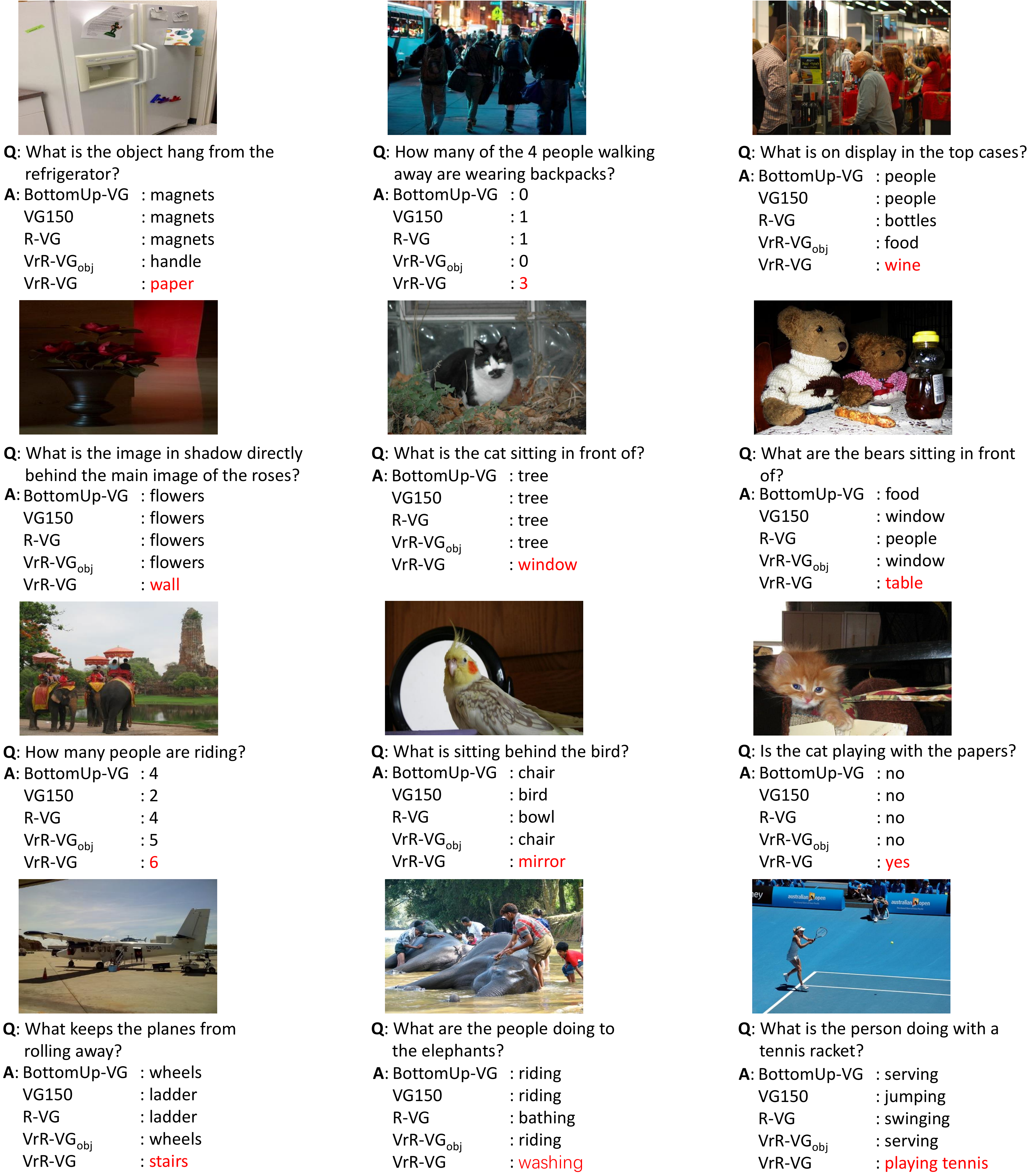}
    \caption{More examples of VQA results.}
    \label{vqa}
\end{figure*}

Additional VQA results are shown in Fig.~\ref{vqa}. Thanks to the relationship-aware representation learned from VrR-VG, most of the hard questions about the interactions of instances can be well answered. 

\FloatBarrier

\clearpage

\section{Results Comparison on Image Captioning Task}

\FloatBarrier

\begin{figure*}[htbp!]
    \centering
    \includegraphics[page=1, width=\textwidth]{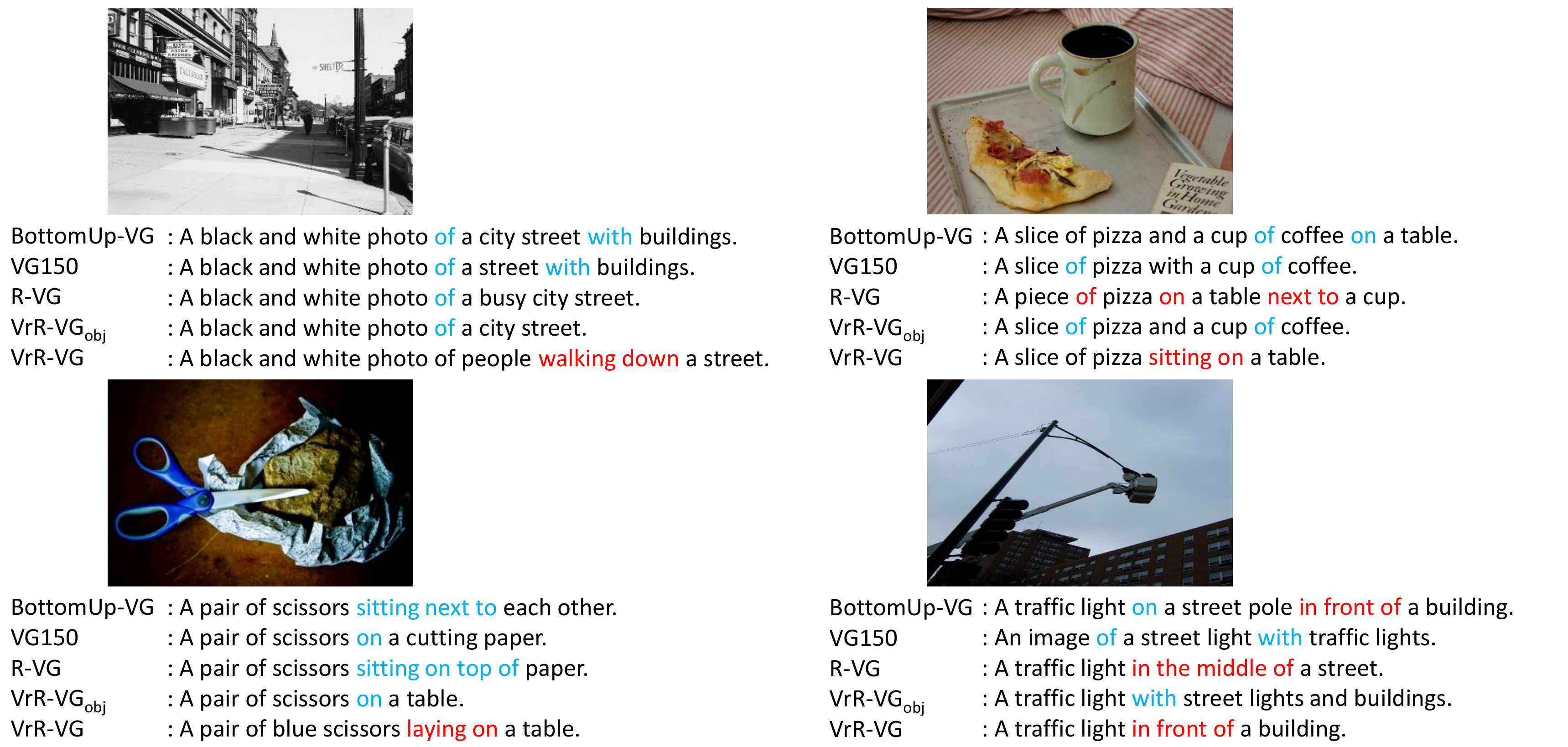}
    \caption{Examples for captioning. The features trained on VrR-VG tends to provide more diverse and vivid expressions in captioning.}
    \label{app_cap_fig}
\end{figure*}

As shown in Fig.~\ref{app_cap_fig}, results in VrR-VG provide more diverse and informative predicates in describing scenes. Although \textit{some expressions like ``scissors laying on a table" may not lead to high scores in captioning metrics}, these vivid and specific results are valuable for cognitive tasks.

\begin{figure*}[htbp!]
    \centering
    \includegraphics[page=2, width=\textwidth]{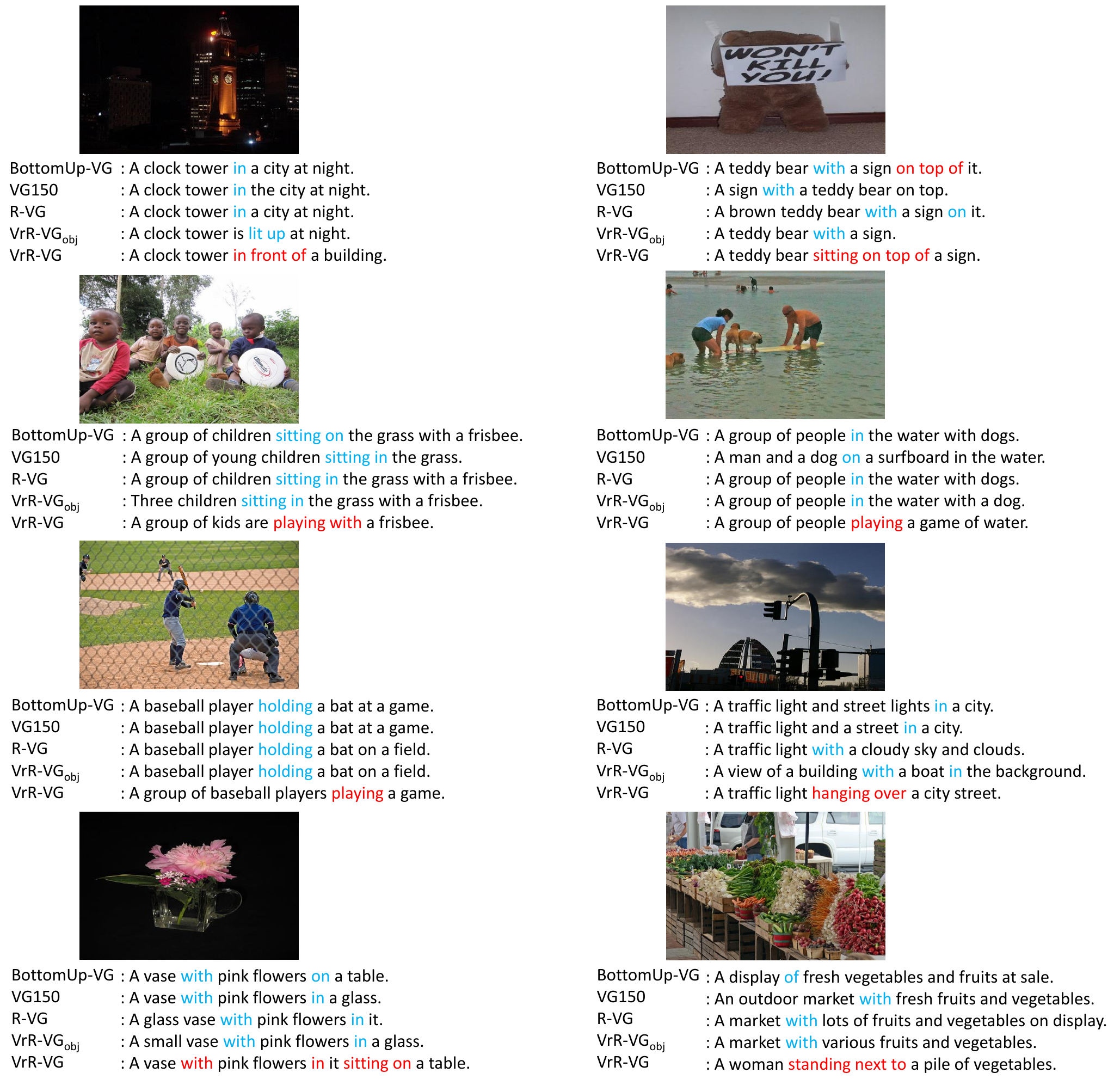}
\end{figure*}

%\fi

\end{document}